\documentclass[letterpaper]{article} 
\usepackage{aaai2027}  
\usepackage[hyphens]{url}  
\usepackage{graphicx} 
\usepackage{natbib}  
\usepackage{caption} 
\usepackage{algorithm}
\usepackage{algorithmic}
\usepackage{amsmath}
\usepackage{adjustbox}
\usepackage{amssymb}
\usepackage{booktabs}
\usepackage{multirow}

\usepackage{colortbl}  
\usepackage{xcolor}   
\usepackage{array}   
\usepackage{pifont}
\usepackage{tabularx}
\usepackage{makecell}

\definecolor{bestgreen}{RGB}{46, 204, 113}
\definecolor{secondgreen}{RGB}{178, 235, 207}
\definecolor{mutedgray}{RGB}{125, 125, 125}

\newcommand{\best}[1]{\textbf{#1}}
\newcommand{\second}[1]{\underline{#1}}

\newcommand{\checkmarkgreen}{\textcolor{green}{\checkmark}}
\newcommand{\cross}{\textcolor{red}{\ding{55}}}

\usepackage{newfloat}
\usepackage{listings}
\DeclareCaptionStyle{ruled}{labelfont=normalfont,labelsep=colon,strut=off} 
\floatstyle{ruled}
\newfloat{listing}{tb}{lst}{}
\floatname{listing}{Listing}

\usepackage{booktabs}

\title{CA-World: Multi-Object Counterfactual Alignment for Efficient Interactive-Ready Reconstruction}
\author {
    Xin Dong\textsuperscript{\rm 1,\rm 2},
    Weijian Deng\textsuperscript{\rm 1},
    Lihan Zhang\textsuperscript{\rm 1},
    Tianru Dai\textsuperscript{\rm 1},
    Wenfeng Deng\textsuperscript{\rm 2},
    Yansong Tang\textsuperscript{\rm 1}\corresponding
}
\affiliations {
    \textsuperscript{\rm 1}Shenzhen International Graduate School, Tsinghua University\\
    \textsuperscript{\rm 2}Pengcheng Laboratory\\
    dong-x23@mails.tsinghua.edu.cn
}

\begin{document}

\maketitle

\begin{abstract}
Reconstructing interaction-ready 3D worlds is essential for physical simulation, virtual reality, robotics, and autonomous driving. However, existing methods mainly optimize static and holistic visual fidelity, with limited support for multi-object interaction. We argue that an interaction-ready reconstruction should anticipate potential scene changes and preserve geometric completeness, visual quality, multi-object spatial relationship, and physical plausibility under potential interactions. To this end, motivated by the causal intervention, we propose CA-World, an efficient framework that integrates counterfactual alignment learning into a decoupling–reintegration reconstruction pipeline. Specifically, we formulate foreground–background decoupling as a visual intervention, separate object generation and background inpainting as counterfactual generation, and scene reintegration as an inverse intervention. According to counterfactual consistency, reversing the intervention should recover the factual world, motivating three alignment objectives between the reintegrated and original scenes: appearance, spatial, and physical consistency. This formulates interaction-ready reconstruction as counterfactual alignment learning with direct supervision. Moreover, leveraging the locality of object-level interventions, CA-World constrains counterfactual states using the observed scene, enabling efficient and coherent reintegration without jointly optimizing all object states, thereby reducing computational cost and error accumulation. Experiments on object completeness, spatial accuracy, outdoor background completion, rendering quality, simulated dynamics, and downstream applications demonstrate the effectiveness of CA-World. Project page: https://chnxindong.github.io/ca-world/.
\end{abstract}

\begin{table*}[t]
\centering
\renewcommand{\arraystretch}{1.3}
\small 
\begin{tabularx}{0.9\linewidth}{>{\centering\arraybackslash}p{3.5cm}*{5}{|>{\centering\arraybackslash}X}}
\toprule
\textbf{\makecell{Method}} & 
\textbf{\makecell{Amodal \\ 3D Recon}} & 
\textbf{\makecell{Background \\ Fidelity}} & 
\textbf{\makecell{Physics \\ Capacity}} & 
\textbf{\makecell{Spatial \\ Accuracy}} & 
\textbf{\makecell{Optimization \\ Efficiency}} \\
\midrule
Amodal3R       & \checkmarkgreen & \cross & \cross & \cross & \cross \\
O2Recon & \checkmarkgreen & \cross & \cross & \cross & \cross \\
Gen3DSR    & \checkmarkgreen & \cross & \cross & \cross & \cross \\
PhysGaussian     & \cross & \checkmarkgreen & \cross & \cross & \checkmarkgreen \\
DecoupledGaussian & \checkmarkgreen & \checkmarkgreen & \cross & \cross & \cross \\
PhyRecon   & \checkmarkgreen & \checkmarkgreen & \checkmarkgreen & \cross & \cross \\
DP-Recon   & \checkmarkgreen & \checkmarkgreen & \cross & \cross & \cross \\
HoloScene & \checkmarkgreen & \checkmarkgreen & \checkmarkgreen & \cross & \cross \\
WorldAct  & \checkmarkgreen & \checkmarkgreen & \cross & \cross & \checkmarkgreen \\
TelePhysics & \checkmarkgreen & \checkmarkgreen & \cross & \checkmarkgreen & \checkmarkgreen \\
SimFoundry & \checkmarkgreen & \checkmarkgreen & \checkmarkgreen & \cross & \checkmarkgreen \\
Robosnap  & \checkmarkgreen & \checkmarkgreen & \checkmarkgreen & \cross & \checkmarkgreen \\
CA-World (Ours)  & \checkmarkgreen & \checkmarkgreen & \checkmarkgreen & \checkmarkgreen & \checkmarkgreen \\
\bottomrule
\end{tabularx}
\caption{Comparison of Interactive 3D World Models. Existing methods fail to simultaneously meet all five interactive-ready reconstruction requirements.}
\label{tab:method_comparison}   
\end{table*}

\section{Introduction}

Interactive-ready reconstruction in 3D environments is important for many applications, including virtual reality~\cite{vr-gs,live-gs}, robotic manipulation~\cite{manigaussian,robogs,robo_twin}, interactive video generation~\cite{bruce2024genie,geng2025motion_prompting,li2024gid}, multi-modal application~\cite{dynamic-pgsr,mf-mod} and world modeling for embodied intelligence~\cite{yu2025wonderworld,physgen,physctrl,animatediff}. Recent advances in neural rendering and Gaussian splatting enable detailed reconstruction of real-world scenes from multi-view images, producing visually accurate environments that support tasks such as novel view synthesis and scene understanding. However, enabling reliable physics-based interaction in reconstructed scenes remains challenging, particularly in real-world environments containing multiple interacting objects.

A fundamental challenge lies in the fact that, prior to potential interactions exerting their influence on a scene, the resulting modifications to objects and backgrounds remain indeterminate.Consequently, a reconstruction that merely describes the factual state may fail to remain geometrically complete, visually coherent, and physically plausible after interaction. Recent methods attempt to address this problem from different perspectives. Generative 3D approaches~\cite{chen2025sam3d,wu2025amodal3r,xiang2025structured} exploit large-scale shape priors to infer plausible object geometry from partial observations, improving geometric completeness under manipulation. However, they cannot guarantee spatial consistency and physical plausibility. Other works bridge holistic reconstruction and physical interaction by representing scene elements as simulation particles~\cite{physgaussian,physdreamer,physflow,omniphysgs}, enabling physically plausible dynamics. Nevertheless, they typically assume relatively simple environments or focus on single-object interactions, and therefore struggle with cluttered real-world scenes multiple objects and mutual occlusions. More recently, HoloScene~\cite{xia2026holoscene} and WorldAct~\cite{hu2026worldact} adopt a decoupling–reintegration paradigm, using segmentation~\cite{groundedsam,ravi2024sam2} and inpainting models to separate foreground objects from the background before recomposing them for joint optimization. Although conceptually straightforward, this formulation requires coupled optimization over all object states, resulting in difficult convergence, substantial computational overhead, and potential error accumulation.

In this work, we emphasize that interaction-ready reconstruction should not merely represent the observed scene, but also anticipate how it may change under potential interactions while preserving geometric completeness, visual fidelity, spatial coherence, and physical plausibility. Motivated by causal intervention, we formulate this problem as counterfactual alignment learning and introduce three counterfactual consistency objectives to provide explicit reconstruction supervision. Specifically, we propose CA-World, an efficient framework that incorporates counterfactual alignment into a decoupling–reintegration pipeline. We interpret foreground–background decoupling as a visual intervention, object generation and background inpainting as counterfactual generation, and scene reintegration as an inverse intervention. Counterfactual consistency requires the inverse intervention to recover the factual world, leading to three complementary objectives: appearance, spatial, and physical consistency. Appearance consistency uses geometry-derived renderings from 3D generative models, such as depth and normal maps, as structural anchors to align the reintegrated counterfactual scene with the observed multi-view images. Spatial consistency injects fine-grained metric scale information from the reconstructed scene to align generated objects with the reconstruction coordinate system, ensuring that factual and counterfactual states share a common spatial reference. Physical consistency further models support and relative-position relationships among objects and between objects and the background. We infer these relationships using a large multimodal model and refine them with metric constraints to improve physical placement accuracy. By exploiting the locality of object-level interventions, CA-World constrains each counterfactual state with the observed scene, enabling coherent and efficient reintegration without jointly optimizing all object states, thereby reducing computational cost and error accumulation. Finally, we employ a Material Point Method (MPM) simulator to model state transitions in the reconstructed world.

Our contributions can be summarized as follows:
\begin{itemize}
\item  We formulate interaction-ready reconstruction for multi-object scenes from a counterfactual perspective, and propose an efficient counterfactual alignment framework built upon a decoupling-reintegration pipeline.

\item We introduce three consistency objectives for counterfactual alignment: appearance consistency aligns generated geometry with multi-view observations, spatial consistency aligns generated objects to the reconstructed evident scene, and physical consistency reinforces inter-object and object--background support relationships. 

\item Extensive experiments demonstrate the effectiveness of CA-World in reconstruction, simulation, and downstream interactive applications.
\end{itemize}

\begin{figure*}[t]
    \centering
    \includegraphics[width=\linewidth]{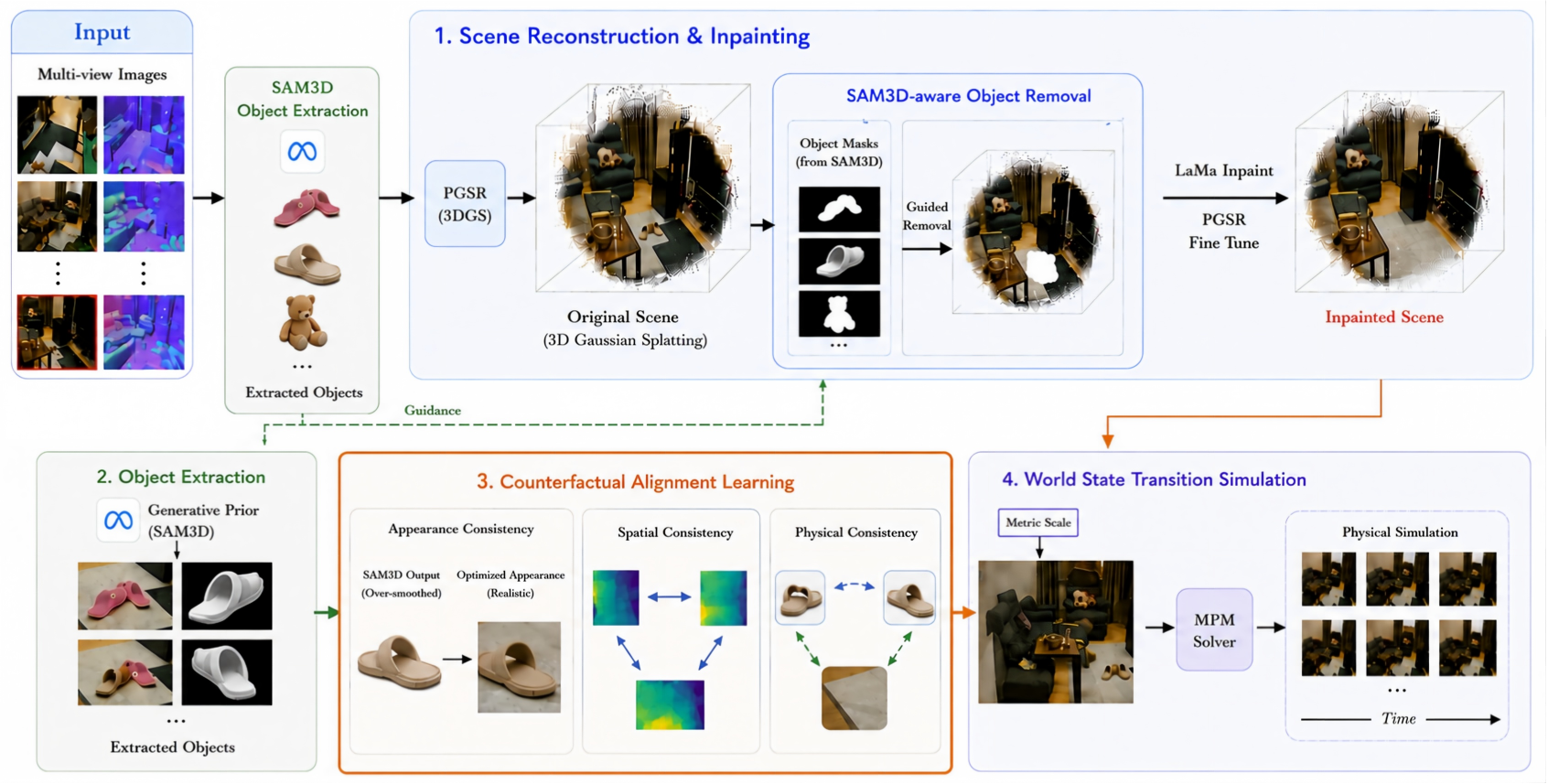}
    \caption{
    CA-World Overview. The pipeline consists of four major steps:
(A) Scene reconstruction, where the scene is reconstructed from multi-view images using PGSR, followed by object removal and inpainting to obtain a clean background scene;
(B) Object extraction, where target objects are segmented and converted into complete 3D geometry using image-to-3D generation with SAM3D;
(C) Counterfactual Alignment, where appearance, spatial, physical consistency objectives to restore the objects’ pose and appearance within the scene;
(D) World State Transition Simulation, where the reconstructed scene is applied simulation using MPM solver to enable interactive dynamics.
    }
    \label{fig:framework_tianru_2}
\end{figure*}

\section{Related Work}

\subsection{Physics-Based Intractive 3D Scene Model}
Physics-based intractive 3D scene model integrates neural representations with physical simulation to model realistic object motion. 
PhysGaussian \cite{physgaussian} embeds Newtonian dynamics into 3D Gaussian representations to model physically plausible deformation and stress without explicit meshes. 
PhysDreamer \cite{physdreamer} distills motion priors from video generative models to enable dynamic responses of static 3D objects.
Learning realistic material properties remains challenging due to limited supervision. 
DreamPhysics \cite{dreamphysics} distills motion priors from video diffusion models to learn material fields that drive physics-based MPM simulations. 
Physics3D \cite{physics3d} extends this direction by learning diverse material properties from diffusion priors and incorporating them into a viscoelastic simulation framework. 
PhysFlow \cite{physflow} further combines multimodal foundation models with video diffusion to refine material parameters for dynamic scene simulation. 
Feature Splatting \cite{feature_splatting} integrates physics simulation with vision–language semantics for automatic material assignment.
DecoupledGaussian \cite{wang2025decoupledgaussian} enables object-level simulation by separating foreground objects from contacted surfaces in real videos. However, it mainly handles single-object separation and assumes planar contact surfaces, limiting multi-object interactions.
HoloScene~\cite{xia2026holoscene} and WorldAct~\cite{hu2026worldact} decouple and reintegrate foreground objects with the background for all object state optimization. However, this approach struggles with convergence and incurs high computational overhead. PhyRecon~\cite{ni2024phyrecon} advances instance-level scene reconstruction by integrating a differentiable physics loss to correct object instability. DP-Recon~\cite{ni2025dprecon} addresses amodal sparse-view reconstruction through the 2D diffusion priors. SimFoundry~\cite{ranawaka2026simfoundrymodularautomatedscene} and RoboSnap~\cite{zhang2026robosnaponeshotrealtosimscene} have comparable decoupling-reintegration pipeline but with differences in task setup.

\subsection{Object Extraction from Images}
Generative models have made significant strides in reconstructing 3D scenes from 2D images. LRM \cite{hong2023lrm} and LGM \cite{tang2024lgm} pioneered the use of feed-forward models to enable rapid inference from single images. To further enhance generation quality, Trellis \cite{xiang2024structured}, Seed3D \cite{feng2025seed3d10imageshighfidelity}, and Hunyuan3D 2.0 \cite{hunyuan3d22025tencent} integrated massive datasets and diffusion priors, achieving high-fidelity results. However, these approaches generally reconstruct the entire image content indiscriminately, often producing geometries that lack the physical realism required for simulation.
For masked object extraction, O$^2$-Recon \cite{hu20242} and Amodal3R \cite{wu2025amodal3ramodal3dreconstruction} and Gen3DSR~\cite{ardelean2025gen3dsrgeneralizable3dscene} proposed utilizing diffusion priors to recover occluded shapes, though their performance was constrained by reliance on synthetic training data. Subsequently, SAM3D \cite{chen2025sam3d} addressed this limitation by leveraging large-scale data to improve robustness in diverse scenarios. Despite these advances, existing works fail to simultaneously extract multiple objects from real scenes and restore them to their original 3D spatial positions. This loss of positional context, combined with insufficient geometric fidelity, renders current pipelines inadequate for downstream physics-based simulations.

\subsection{Object Pose Estimation}

Object 6DoF pose estimation aims to recover the 3D position and orientation of objects from visual observations. 
Existing methods are broadly categorized into model-based and model-free approaches.
Model-based methods assume access to object geometry during inference. 
MegaPose \cite{megapose} estimates poses of unseen objects using large-scale synthetic training data and a render-and-compare refinement strategy. 
GS-Pose \cite{gspose} constructs multiple object representations from posed RGB images and refines poses using differentiable 3D Gaussian Splatting rendering. 
Pos3R \cite{pos3r} leverages 3D reconstruction foundation models to extract geometry-consistent features, enabling training-free pose estimation from a single RGB image.
Model-free approaches estimate object poses without object models. 
iG-6DoF \cite{ig6dof} proposes an iterative framework based on 3D Gaussian Splatting that generates initial pose hypotheses using rotation-equivariant features and refines them through render-and-compare optimization.
Despite these advances, robust pose estimation in complex scenes with large motions and multiple interactions remains challenging.

\section{Approach}

\subsection{Overall Pipeline}
Given multi-view images $\{I_0, I_1, ...\}$ and their camera parameters $\{p_0, p_1, ...\}$, interaction-ready reconstruction aims to recover an object-centric 3D world that faithfully explains the observations while remaining geometrically complete, spatially coherent, visually realistic, and physically plausible. As illustrated in Fig.\ref{fig:framework_tianru_2}, our framework consists of four stages:
First, we reconstruct the 3D scene from multi-view images using PGSR\cite{pgsr} together with segmentation maps generated by SAM~\cite{ravi2024sam2}. We then remove the user-prompted objects, inpaint the resulting background holes using LaMa~\cite{suvorov2021lama}, and reconstructed the background to obtain evident scene.
Second, we perform object extraction using generative 3D priors. Specifically, we leverage SAM3D~\cite{chen2025sam3d} to recover complete object geometry from partial observations. By formulating multi-object decoupling as a decoupled object-generation problem, SAM3D enables the recovery of complete 3D shapes for individual objects that are originally coupled within the reconstructed scene.
Third, based on the principle of counterfactual consistency which states that reversing an intervention should recover the factual world, we propose three alignment objectives, namely appearance, spatial, and physical consistency, between the reintegrated and original scenes. These objectives facilitate the transition from interaction-ready reconstruction to counterfactual alignment learning, ensuring that the intervention can be reliably reversed to reconstruct the factual world.
Finally, the reconstructed scene is applied world state transition using a Material Point Method (MPM) solver, enabling multi-object physical interactions. 

\subsection{Counterfactual Alignment Learning}

\begin{figure}[t]
    \centering
    \includegraphics[width=\linewidth]{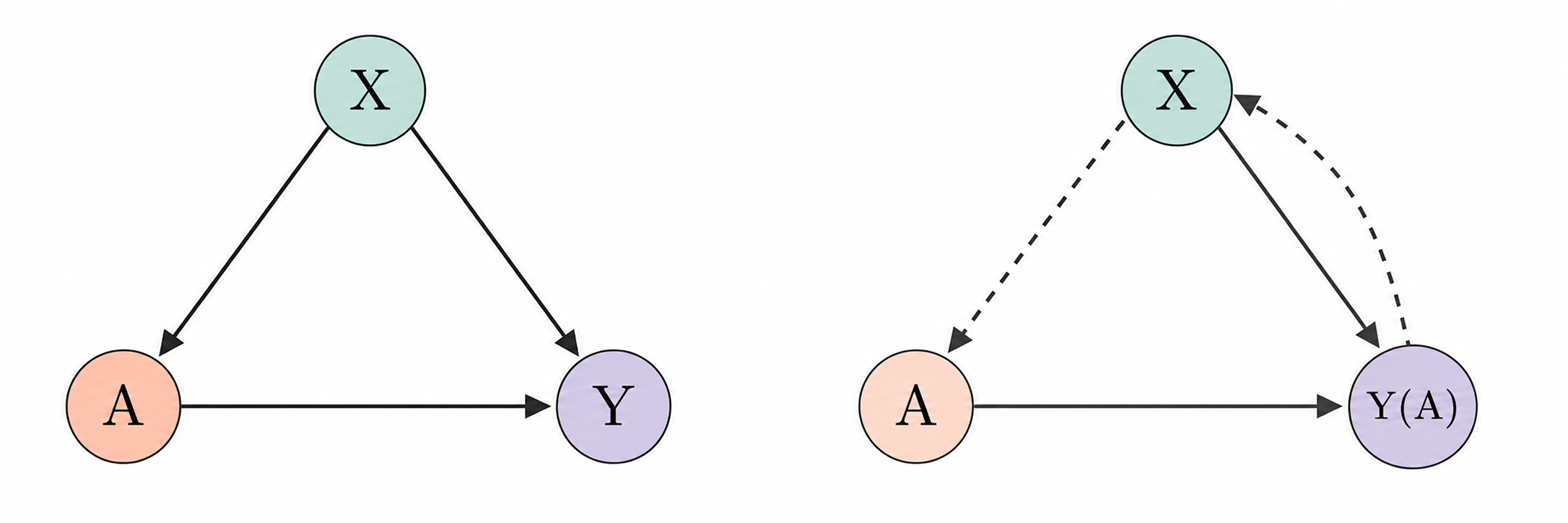}
    \caption{
     Causal intervention graph. We formulate foreground-background decoupling as a visual intervention, object generation and background inpainting as counterfactual generation, and scene reintegration as an inverse intervention. Following the principle of counterfactual consistency, we formulate interaction-ready reconstruction as directly supervised counterfactual alignment learning.
}
    \vspace{-5pt}
    \label{fig:ci}
\end{figure}

We argue that interaction-ready reconstruction must anticipate scene changes while preserving geometric, visual, spatial, and physical integrity. As shown in Fig.~\ref{fig:ci}, we formulate foreground-background decoupling as a visual intervention (where the transition from the factual visual world X to the visual intervention object A is represented by a dashed line), object generation and background inpainting as counterfactual generation ($Y(A)$), and scene reintegration as an inverse intervention (indicated by a dashed arrow from $Y(A)$ back to A). Guided by the principle of counterfactual consistency, which requires the inverse operation to recover the factual world, we introduce appearance, spatial, and physical alignment objectives. This casts interaction-ready reconstruction as directly supervised counterfactual alignment learning.

\paragraph{Appearance Consistency.} 
Decoupled object-generation modeling can effectively address object incompleteness in 3D reconstruction caused by occlusions, and thus preserves accurate geometry for separated objects. However, it does not reliably maintain appearance consistency. As shown in Fig.\ref{fig:framework_tianru_2}, objects exhibit noticeable cartoon-like artifacts before appearance alignment. To address this, we render the SAM3D generated objects for geometric cues (depth and normal map), and apply a patch-level VGG feature loss\cite{simonyan2014vgg}  within visible region. Owing to the property of 3DGS that local-region supervision can optimize particle attributes over a broader spatial extent~\cite{3dgs}, this optimization does not produce hard boundaries across mask borders.

\paragraph{Spatial Consistency.} 
To ensure that each separated object preserves accurate spatial placement, we perform render-and-compare refinement with spatial metric exploitation. The render-and-compare refinement module iteratively optimizes each object’s pose ($i.e.$, translation and rotation) by comparing rendered masked-object images with the corresponding ground-truth observations. To further enhance optimization, we exploit metric cues from 3D reconstruction ($e.g.$, depth maps) to compute pointmaps, which provide SAM3D with more accurate 3D initialization. Specifically, given a depth map $D \in \mathbb{R}^{H \times W}$, camera intrinsics $f_x, f_y, c_x, c_y$, we obtain the pointmap $\mathbf{P}$ via pinhole-camera back-projection:
\begin{equation}
\mathbf{P}(v, u) = 
\begin{bmatrix}
X \\ Y \\ Z
\end{bmatrix}
=
\begin{bmatrix}
\dfrac{(u - c_x) \cdot D(v, u)}{f_x} \\[10pt]
\dfrac{(v - c_y) \cdot D(v, u)}{f_y} \\[10pt]
D(v, u)
\end{bmatrix}
\end{equation}
The resulting point map is an array of shape $(H, W, 3)$, where each pixel stores its 3D coordinate $(X,Y,Z)$ in the camera coordinate system. For pose refinement, we separately update translation and rotation. Following \cite{ig6dof}, this process is formulated as: 

\begin{equation}
\begin{array} { l } { { \displaystyle t _ { \Delta } ^ { k + 1 } = \arg \operatorname* { m i n } _ { t _ { \Delta } ^ { k + 1 } } \mathcal { L } _ { t } ( G_{Render} ( t _ { \Delta } ^ { k + 1 } + t ^ { k } , \mathcal{O} ) , I _ {gt} ) } } \\ { { \displaystyle + \arg \operatorname* { m i n } _ { R _ { \Delta } ^ { k + 1 } } \mathcal { L } _ { R } ( G_{Render} ( R _ { \Delta } ^ { k + 1 } \odot ( t _ { \Delta } ^ { k + 1 } + t ^ { k } ) , \mathcal{O} ) , I _ {gt} ) , } } \end{array}
\end{equation}
By minimizing the discrepancy between the rendered mask and the ground-truth image, we can progressively refine the object's spatial placement, ensuring it aligns well with its original scene context. 

\begin{table*}[t]
\centering

\caption{
Evaluating geometric completeness and spatial relationships on Replica, ScanNet++, and iGibson.
The best and second-best results are bold and underlined, respectively.
}

\label{tab:primary_table}

\fontsize{11pt}{12pt}\selectfont

\setlength{\tabcolsep}{1.8pt}
\renewcommand{\arraystretch}{1.1}

\begin{tabular*}{0.9\textwidth}{@{\extracolsep{\fill}}
c l ccc ccc ccc}

\toprule

&
\multirow{2}{*}{\textbf{Method}}
&
\multicolumn{3}{c}{\textbf{Geometry}}
&
\multicolumn{3}{c}{\textbf{Rendering}}
&
\multicolumn{3}{c}{\textbf{Spatial Relation}}
\\[-1mm]

\cmidrule(lr){3-5}
\cmidrule(lr){6-8}
\cmidrule(lr){9-11}

&
& \rule{0pt}{3ex}
CD$\downarrow$
& F1$\uparrow$
& NC$\uparrow$
& PSNR$\uparrow$
& SSIM$\uparrow$
& LPIPS$\downarrow$
& OR\%$\uparrow$
& \makecell{Stable\\(Ground)\%$\uparrow$}
& \makecell{Stable\\(All)\%$\uparrow$}
\\

\midrule
\addlinespace[2pt]


\multirow{4}{*}{\rotatebox[origin=c]{90}{Replica}}

& PhyRecon
& 4.52
& 71.07
& 92.06
& 23.19
& 0.764
& 0.434
& 77.5
& 56.5
& 5.6
\\

& DP-Recon
& \second{3.45}
& \second{87.66}
& \second{94.23}
& 22.10
& 0.728
& 0.420
& 56.3
& 21.7
& 8.5
\\

& Holoscene
& 4.05
& 83.21
& 92.21
& \second{27.82}
& \second{0.849}
& \second{0.304}
& \best{100.0}
& \second{95.7}
& \second{81.7}
\\

& Ours
& \best{3.25}
& \best{88.32}
& \best{95.52}
& \best{28.32}
& \best{0.870}
& \best{0.261}
& \best{100.0}
& \best{96.3}
& \best{83.3}
\\

\addlinespace[4pt]
\midrule
\addlinespace[4pt]


\multirow{4}{*}{\rotatebox[origin=c]{90}{ScanNet++}}

& PhyRecon
& 31.16
& 39.57
& 82.28
& 22.32
& 0.791
& 0.432
& 92.9
& 67.3
& 9.4
\\

& DP-Recon
& 22.96
& \second{65.48}
& 87.13
& 21.44
& 0.715
& 0.466
& 90.6
& 20.0
& 9.4
\\

& Holoscene
& \second{21.93}
& 63.11
& \second{88.09}
& \second{25.88}
& \second{0.873}
& \second{0.268}
& \best{100.0}
& \second{93.9}
& \second{70.6}
\\

& Ours
& \best{20.33}
& \best{66.23}
& \best{89.02}
& \best{26.91}
& \best{0.891}
& \best{0.238}
& \best{100.0}
& \best{94.2}
& \best{72.8}
\\

\addlinespace[4pt]
\midrule
\addlinespace[4pt]


\multirow{4}{*}{\rotatebox[origin=c]{90}{iGibson}}

& PhyRecon
& \second{11.27}
& \second{45.49}
& \second{83.85}
& \second{27.40}
& \second{0.860}
& 0.333
& 62.9
& 45.3
& 5.2
\\

& DP-Recon
& 30.31
& 21.89
& 70.81
& 21.94
& 0.728
& 0.432
& 74.2
& 16.3
& 4.1
\\

& Holoscene
& 12.00
& 34.15
& 82.91
& 25.88
& 0.854
& \second{0.301}
& \best{100.0}
& \best{74.4}
& \best{71.1}
\\

& Ours
& \best{11.12}
& \best{47.37}
& \best{86.26}
& \best{28.21}
& \best{0.882}
& \best{0.291}
& \best{100.0}
& \second{73.8}
& \second{70.9}
\\

\bottomrule

\end{tabular*}

\end{table*}

\paragraph{Physical Consistency.} 
Motivated by \cite{li2025scorehoi}, we therefore introduce a physics-constrained consistency alignment, which augments reconstruction with both scene-object and object-object physical constraints. Specifically, we use a pair of slippers as an example: the initial alignment exhibits severe interpenetration, which is physically implausible. We feed the image into a large multimodal model ($i.e.$ Gemini 3 Pro) to infer a relation graph describing scene-object and object-object interactions. Based on this relation-control graph, we define two physical constraints: 

1) object-scene relation constraint, which ensures the object maintains stable contact with the ground without floating or sinking. This can be written as:
\begin{equation}
\begin{aligned}
\mathcal{L}_{\text{os}}
=
\frac{1}{|\mathcal{S}_o|}
\sum_{\mathbf{x}_i\in\mathcal{S}_o}
\left[
\max\!\big(0,\ \mathbf{n}^\top\mathbf{x}_i+d-\epsilon\big)^2
\right. \\
\left.
+ \max\!\big(0, -(\mathbf{n}^\top\mathbf{x}_i+d)-\epsilon\big)^2
\right].
\end{aligned}
\end{equation}
2) object-object non-penetration constraint, which prevents interpenetration between objects by enforcing a positive minimum distance between them. This process can be written as:
\begin{equation}
\mathcal{L}_{\text{oo}}
=
\sum_{(a,b)\in\mathcal{E}_{oo}}
\Big[\max\!\big(0,\; m-d_{\min}(\mathcal{O}_a,\mathcal{O}_b)\big)\Big]^2,
\end{equation}
\begin{equation}
d_{\min}(\mathcal{O}_a,\mathcal{O}_b)
=
\min_{\mathbf{x}\in\mathcal{O}_a,\ \mathbf{y}\in\mathcal{O}_b}
\|\mathbf{x}-\mathbf{y}\|_2.
\end{equation}
After applying our physical consistency objective, the slippers are separated to a physically reasonable distance, eliminating interference during simulation.

\subsection{World State Transition Simulation}
After the interactive reconstruction, we can apply world state transition using a Material Point Method algorithm (MPM)~\cite{hu2018mls-mpm}. 
The Material Point Method (MPM) algorithm synergistically combines the advantages of both Lagrangian and Eulerian frameworks by representing the continuum material as a collection of particles, denoted as $\mathcal{P}_{\text{MPM}} = \{(\mathbf{x}_p, \mathbf{v}_p, \mathbf{F}_p)\}$, where each particle encapsulates a localized material volume. These particles carry intrinsic state variables, including position \(\mathbf{x}_p\), velocity \(\mathbf{v}_p\), and deformation gradient \(\mathbf{F}_p\). The Lagrangian nature of the particles inherently guarantees mass conservation, whereas the auxiliary Eulerian background grid enables robust enforcement of momentum conservation. The interaction between particles and the grid is mediated through B-spline kernel functions, which facilitate accurate and smooth transfer of field quantities. Momentum conservation is enforced discretely in time: at each time step, particle momenta are mapped to the grid, grid velocities are updated according to the governing equations of motion, and the resulting velocities are then interpolated back to the particles to adjust their positions:
\begin{equation}
    \mathbf{x}_p^{t+1} = \mathbf{x}_p^t + \Delta t \mathbf{v}_p^{t+1}.
\end{equation}
The deformation gradient $\mathbf{F}_p$ is updated incrementally, with plasticity corrections, enabling realistic simulation of complex deformations and interactions in dynamic scenes. For the interactive forces, users can define impulse force at point and force field, which support more flexible interactions. The interactive simulation can run on a consumer GPU (NVIDIA RTX 4090). Details on MPM can be found in supplementary material.

\section{Experiments}

\begin{figure*}[t]
    \centering
    \includegraphics[width=1.0\linewidth]{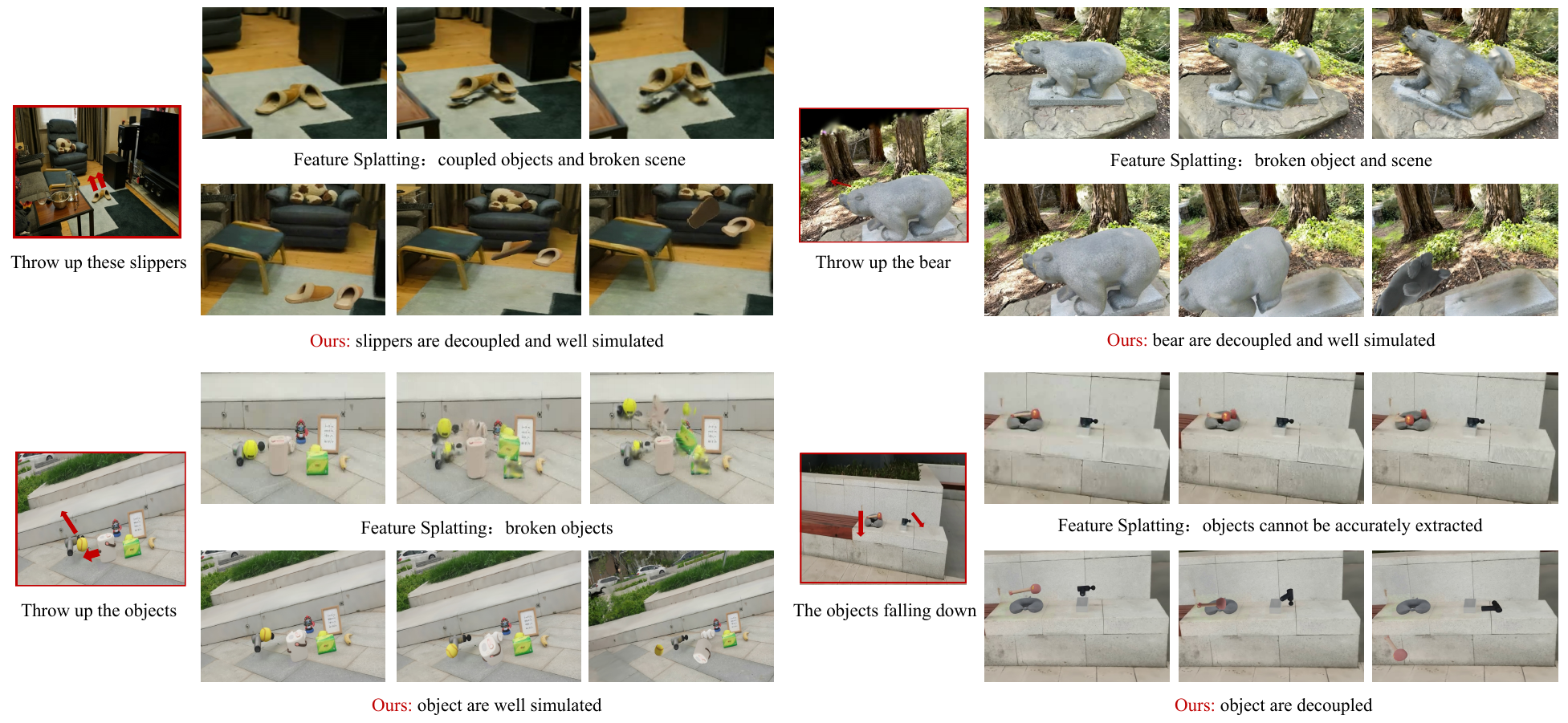}
    \caption{Comparisons with the state of the arts. Our method ensures accurate decoupling, alignment, and fidelity. Feature Splatting suffers from fragmentation due to poor background separation. }
    \label{fig:compare_new_3}
\end{figure*}

\subsection{Experimental Settings}

\paragraph{Datasets.} To evaluate geometric completeness and spatial relationships, following Holoscene~\cite{xia2026holoscene}, we utilized three scenes from Replica~\cite{replica19arxiv}, three from ScanNet++~\cite{yeshwanth2023scannet++}, and two from iGibson~\cite{shen2021igibson}. For visual quality and interactive simulation, we selected two real-world static scenes (bear and room) from DecoupledGaussian~\cite{wang2025decoupledgaussian} and captured two multi-object coupled scenes (outdoor and flower bed) using a mobile phone. 

\paragraph{Metrics.} Geometric completeness is evaluated using Chamfer Distance (CD), F-Score (F1), and Normal Consistency (NC), while rendering quality is assessed via PSNR, SSIM, and LPIPS. Following Holoscene~\cite{xia2026holoscene}, multi-object spatial relationships are evaluated in Isaac Sim~\cite{mittal2023orbitisaacsim} by reporting the stability ratio (Stable\%) and object reconstruction ratios (OR\%). Given the challenges in evaluating simulated dynamics, we report user study and LLM-as-judge results. In ablation studies, we define edge error as the horizontal and vertical Sobel edge error between objects in rendered frames and their ground-truth images, effectively capturing the ablated visual degradation. 

\begin{table}[t]
    \small
    \centering
    \renewcommand{\arraystretch}{1.15}
    \setlength{\tabcolsep}{10pt}
    \caption{User study and LMM-as-Judge evaluation results. Both metrics are higher-is-better. Our method outperforms the baselines in both motion realism and visual quality, demonstrating its effectiveness in generating physically plausible motions while maintaining high visual fidelity. }
    \label{tab:scoring}
    \begin{tabular}{p{0.27\linewidth}cc}
        \toprule
        \textbf{Motion Realism} & \textbf{User Study} & \textbf{LMM-as-Judge Eval.} \\
        \midrule
        Feature Splatting & 16 & 10 \\
        DecoupledGaussian$^\star$ & 28 & 20 \\
        Ours & \textbf{41} & \textbf{43} \\
        \midrule
        \textbf{Visual Quality} & \textbf{User Study} & \textbf{LMM-as-Judge Eval.} \\
        \midrule
        Feature Splatting & 15 & 15 \\
        DecoupledGaussian$^\star$ & 37 & 20 \\
        Ours & \textbf{42} & \textbf{33} \\
        \bottomrule
    \end{tabular}
\end{table}

\paragraph{Compared Methods.} We evaluate simulated frames against Feature Splatting~\cite{feature_splatting} in multi-object interactive simulation and rendering. For a broader quantitative assessment in single-object condition, we additionally include DecoupledGaussian~\cite{wang2025decoupledgaussian}. To evaluate geometric completeness and spatial relationships, we adopt amodal 3D scene reconstruction evaluation, with PhyRecon~\cite{ni2024phyrecon}, DP-Recon~\cite{ni2025dprecon}, and Holoscene~\cite{xia2026holoscene} as baselines. Note that although WorldAct~\cite{hu2026worldact} shares a similar task formulation, its source code is not publicly available. TelePhysics~\cite{zhang2026telephysicsphysicsgroundedmultiobjectscene}, SimFoundry~\cite{ranawaka2026simfoundrymodularautomatedscene}, and RoboSnap~\cite{zhang2026robosnaponeshotrealtosimscene} are excluded for comparisons due to differences in task setup, despite employing a comparable decoupling-reintegration pipeline.

\begin{table}[t]
    \small
    \centering
    \setlength{\tabcolsep}{4pt}
    \caption{Efficiency studies on reconstruction and simulation modules. $^\star$ represents the results of DecoupledGaussian is for single-object condition.}
    \label{tab:efficiency}

    \begin{tabular*}{\linewidth}{@{\extracolsep{\fill}}
                    @{\hspace{12pt}}l
                    >{\centering\arraybackslash}p{0.22\linewidth}
                    >{\centering\arraybackslash}p{0.22\linewidth}}
        \toprule
        \textbf{Method} & 
        \textbf{Peak GPU Memory $\downarrow$} & 
        \textbf{Simulation Time $\downarrow$} \\
        \midrule
        HoloScene & 24GB & N/A \\
        Feature Splatting & 17GB & 600s \\
        DecoupledGaussian$^\star$ & 15GB & 300s \\
        Ours & \textbf{15GB} & \textbf{300s} \\
        \bottomrule
    \end{tabular*}
\end{table}

\begin{table}[t]
    \centering
    \setlength{\tabcolsep}{6pt}
    \caption{Ablation studies on counterfactual alignment. Results show that three consistency objectives contribute to better quantitative performances. A.C. denotes appearance consistency, S.C. denotes spatial consistency, P.C. denotes physical consistency.}
    \label{tab:ablation}
    
    \begin{tabular*}{\linewidth}{@{\extracolsep{\fill}}cccc}
        \toprule
        Module & Edge Error$\downarrow$ & PSNR$\uparrow$ & SSIM$\uparrow$ \\
        \midrule
        Baseline & 1.739 & 23.92 & 0.8693 \\
        + A.C. & 1.716 & 28.86 & \textbf{0.9602} \\
        + S.C. & 1.227 & 29.12 & 0.9482 \\
        + P.C. & \textbf{1.075} & \textbf{30.04} & \underline{0.9537} \\
        \bottomrule
    \end{tabular*}
\end{table}

\subsection{Comparison with the State of the Art}

\paragraph{Geometric Completeness and Multi-Object Relation.} To evaluate geometric completeness and spatial relation, we follow amodal 3D scene reconstruction evaluation and compare geometric completeness after object separation against PhyRecon, DP-Recon and Holoscene. As shown in Table~\ref{tab:primary_table}, our approach achieves superior object separation while preserving geometric fidelity and accurate spatial relationships, which is essential for interactive-ready reconstructions.

\textbf{Interactive Simulation and Rendering.} To validate the counterfactual consistency for interactive simulation, we show comparisons with Feature Splatting~\cite{feature_splatting} in multi-object condition. As shown in Fig.~\ref{fig:compare_new_3}, our method delivers more accurate object decoupling and restoration, with good simulated dynamics. In contrast, Feature Splatting often fails to separate objects from the background, leading to fragmented objects and scene structure. We also include DecoupledGaussian~\cite{wang2025decoupledgaussian} as a single-object baseline (marked with $ \star $ ). For evaluation, 12 participants assessed rendered videos across all scenes. As shown in Table~\ref{tab:scoring}, Large Multimodal Model (LMM, Gemini 3.1 Pro) assessments align closely with human judgments, validating our approach. Furthermore, efficiency comparisons in Table~\ref{tab:efficiency} for the reconstruction and simulation modules demonstrate that our method is both highly efficient and cost-effective.

\subsection{Ablation Studies}

To further validate the contribution of each consistency objective, we conduct ablation studies summarized in Table~\ref{tab:ablation}, employing edge error, PSNR, and SSIM to quantify the ablated visual degradation. The results indicate that appearance consistency is critical for enhancing visual metrics, whereas spatial and physical consistency are vital for preserving geometric fidelity. 
Specifically, optimizing these consistency objectives facilitates counterfactual alignment, thereby enabling precise causal reasoning under potential and unseen interactions.

\begin{figure}[t]
    \centering
    \includegraphics[width=\linewidth]{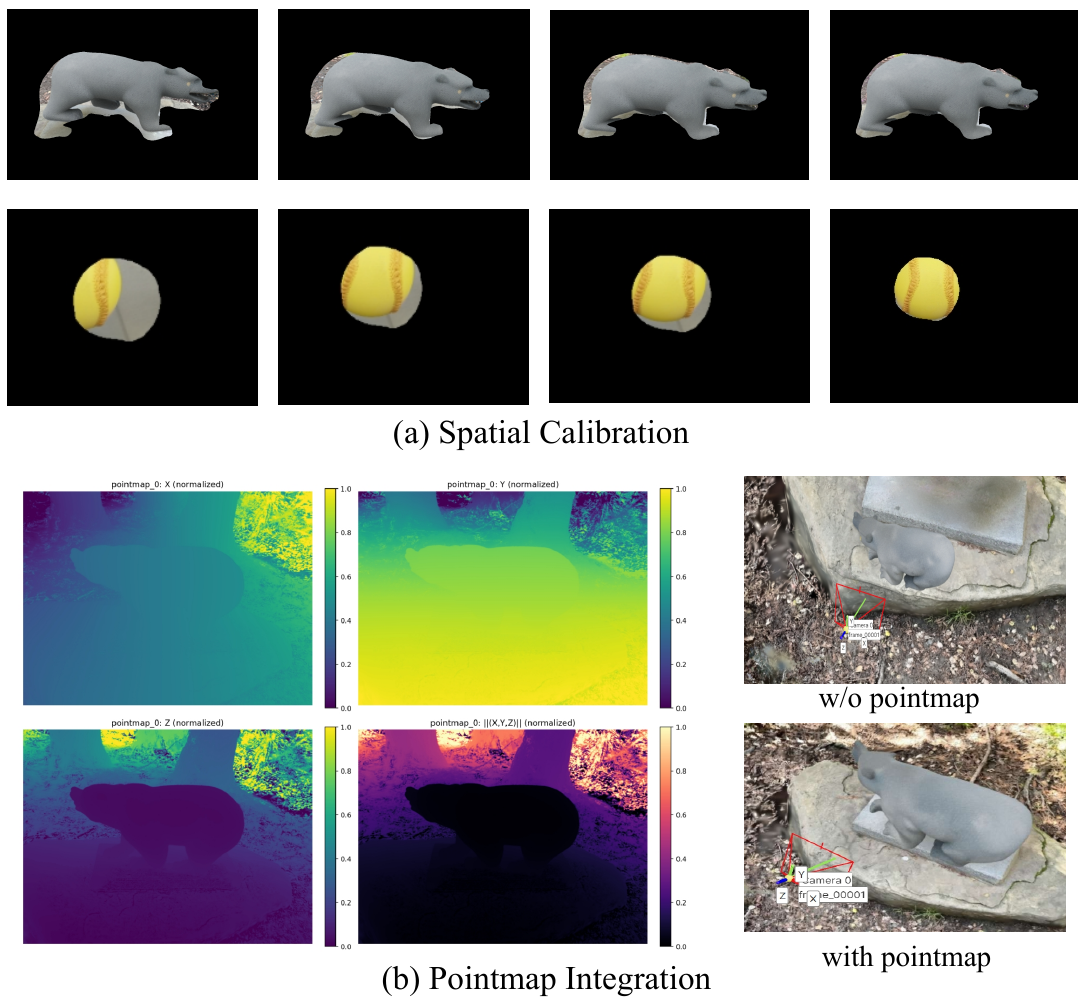}
    \caption{
     Visualization of spatial calibration and pointmap integration process. In sub-figure (a), each column shows the intermediate result at increasing optimization steps. In sub-figure (b), metric cue from pointmap calibrates object to precise position.
}
    \vspace{-5pt}
    \label{fig:pose_alignment_2}
\end{figure}

\textbf{Visualization of Ablations.} Additionally, Fig.~\ref{fig:pose_alignment_2} (a) illustrates the alignment of the object in the bear scene and a ball in the outdoor scene; as optimization progresses, the object masks exhibit increasingly precise alignment with the target regions, and the object positions converge to their correct locations. In Fig.\ref{fig:pose_alignment_2} (b), we ablate spatial consistency, exploiting metric cues from the 3D reconstruction to compute pointmaps, helping 3D generative models with more accurate object initialization in 3D space. These results demonstrate that the consistency optimization introduced by counterfactual alignment can effectively leverage the reconstructed evident scene to reinforce 3D generative states.

\section{Conclusion}

We presented CA-World, an efficient framework for reconstructing interaction-ready 3D worlds. Rather than describing only the current scene state, CA-World anticipates potential object-level changes and preserves geometric completeness, visual coherence, spatial consistency, and physical plausibility under interaction. Motivated by causal intervention, we formulated interaction-ready reconstruction into counterfactual alignment learning through appearance, spatial, and physical consistency objectives. By further exploiting the locality of object-level interventions, CA-World constrains generated states with factual observations, enabling coherent reintegration without jointly optimizing all object states. Extensive experiments demonstrate the effectiveness of CA-World in object reconstruction, spatial placement, background completion, novel-view rendering, physical simulation, and downstream interactive applications. 

\paragraph{Limitations.} Appearance inconsistencies caused by illumination variations, such as specular highlights and shadow changes, remain challenging in scene-centric reconstruction and inpainting. In addition, many scene completion approaches rely on contextual cues, and performance may degrade when the environment provides limited visual context. Addressing these challenges remains an important direction for future research.

\bibliography{reference}


\end{document}